\documentclass[10pt,twocolumn,letterpaper]{article}

\usepackage[pagenumbers]{cvpr} % To force page numbers, e.g. for an arXiv version

\usepackage{graphicx}
\usepackage{amsmath}
\usepackage{amssymb}
\usepackage{booktabs}

\usepackage[pagebackref,breaklinks,colorlinks]{hyperref}

\usepackage[capitalize]{cleveref}
\crefname{section}{Sec.}{Secs.}
\Crefname{section}{Section}{Sections}
\Crefname{table}{Table}{Tables}
\crefname{table}{Tab.}{Tabs.}

\def\cvprPaperID{*****} % *** Enter the CVPR Paper ID here
\def\confName{CVPR}
\def\confYear{2022}

\begin{document}

\title{Stylized Projected GAN: A Novel Architecture for Fast and Realistic Image Generation  }

\author{Md Nurul Muttakin\\
KAUST\\
Thuwal, KSA\\
{\tt\small mdnurul.muttakin@kaust.edu.sa}
\and
Malik Shahid Sultan\\
KAUST\\
Thuwal, KSA\\
{\tt\small malikshahid.sultan@kaust.edu.sa}
\and
Robert Hoehndorf\\
KAUST\\
Thuwal, KSA\\
{\tt\small robert.hoehndorf@kaust.edu.sa}
\and
Hernando Ombao\\
KAUST\\
Thuwal, KSA\\
{\tt\small hernando.ombao@kaust.edu.sa}
}

\maketitle

%%%%%%%%% ABSTRACT
\begin{abstract}

   Generative Adversarial Networks are used for generating the data using a generator and a discriminator, GANs usually produce high-quality images, but training GANs in an adversarial setting is a difficult task. GANs require high computation power and hyper-parameter regularization for converging. Projected GANs tackle the training difficulty of GANs by using transfer learning to project the generated and real samples into a pre-trained feature space. Projected GANs improve the training time and convergence but produce artifacts in the generated images which reduce the quality of the generated samples, we propose an optimized architecture called Stylized Projected GANs which integrates the mapping network of the Style GANs with Skip Layer Excitation of Fast GAN. The integrated modules are incorporated within the generator architecture of the Fast GAN to mitigate the problem of artifacts in the generated images.

\end{abstract}

%%%%%%%%% BODY TEXT
\section{Introduction}

Generative Adversarial Network is a generative deep learning model architecture comprising two parts, the generator which generates the unknown data distribution from a latent space, and a discriminator which takes the input of both the generated samples and real samples and classifies an input instance as either real coming from the real data set or fake coming from the generator, this setting is an adversarial learning setting where the two parts of the network compete against each other having contrasting goals. The goal of the generator is to deceive the discriminator to believe that the generated sample was real on the other hand the objective of the discriminator is to correctly classify both the real and fake samples, if one goal is maximized the other goal is automatically compromised, therefore training GANs is a challenging task, due to this contrasting objective and minimax adversarial game, GANs are notorious to train, and suffer from problems of gradient vanishing, mode collapse, etc., therefore it is difficult to jointly train both parts of the GAN network simultaneously, a moreover significant amount of time and computing resources are required to train the GANs. Researchers have explored the discriminator regularization techniques to improve the balance in the adversarial game including gradient penalties \cite{mescheder2018training} and engineered loss functions, however, it is observed that these strategies are sensitive to the choice of the hyperparameters.GANs \cite{goodfellow2014generative} find their applications in Computer Vision tasks of image translation, realistic photos, and creative art creation \cite{elhoseiny2019creativity}.Various bottlenecks appear in the process of training GANs including computing resources, and data availability, usually GANs require large data sets to generalize and fit the data distribution however in real-life scenarios the data can be a bottleneck in certain disciplines like medical and biology where the underlying data for a specific rare disease might be very small, therefore there is a need to build Generative Models that can quickly learn from small data samples.

Pretrained Networks are widely used in Computer Vision \cite{kolesnikov2020big, ranftl2019towards} , to tailor the learned representations to adapt a specific data set, which reduces the model tuning and building time. Transfer Learning is the process of utilizing the learning parameters from a large pre-trained network whose accuracy is proven and state of the art and was trained on a large similar data-set, for a novel similar data set. Transfer Learning can be used in the cases where the data set available for building a deep learning model is very small\cite{weiss2016survey}.Transfer Learning can be used to speed up the feature extraction and learning in the discriminator part to reduce the training time for the network. Transfer learning is widely used in computer vision to share the learned features from different data sets to be generalized to new data sets with training limited to a few layers for learning data specific features that cannot be learned in the Transfer Learning baseline model, however, tuning the parameters of the resulting model is very critical otherwise this leads to worse results and performance\cite{zhao2020differentiable}. The training time for generating realistic images using Style GAN2\cite{karnewar2020msg}, and Big GAN \cite{brock2018large}, remains very high, therefore there is a need to find a tradeoff between quality and the computation time.Using pre-trained neural networks with GANs produced good results \cite{richter2021enhancing,wang2018high}. \cite{sauer2021projected} investigated the use of pre-trained features to improve the quality of the generated images and stabilize the GANs training. Concatenating the Employing the pre-learned representations directly for training GANs did not improve the training rather resulted in vanishing gradients which lead to the difficulty in training the Generator \cite{arjovsky2017towards} . The researchers explored the learned representations utilization for training GANs by incorporating the feature pyramids and random projections, this resulted in decreasing the computational time required to train the GANs at a lower FID score, however it induced problems relating to the Generated Image Quality like floating heads and artifact \cite{sauer2021projected}. Projected GAN uses a novel architecture for the discriminator however for the generator part the network uses the architecture of the Fast GAN which uses skip layer excitation instead of progressive growing GANs as in Style GAN. We hypothesize that the artifacts produced in the earlier low-resolution layers are transmitted and amplified in the progressive large resolution layers due to up-sampling, therefore if the rate of generating artifacts in the low-resolution layer can be minimized more realistic images can be generated. In this research we propose to combine the prominent advantages of Style GANs and, Projected GANs which are based on a generator of Fast GAN, to solve the problem of generating quality  Images with minimal training, we hypothesize that by using the mapping network similar to style GAN the network can learn to produce high-quality images, by reducing the production of the artifacts in the low-resolution layers, therefore the problem of artifacts and floating heads which is observed in Projected GAN can be resolved, this is novel task due to the reason that there is no single mixing strategy or architecture that can guarantee high-quality images and low training time. This research aims at exploring various architecture designs for the Generator part of the GAN network to achieve the objective of generating quality images from projection space.

%%%%%%%%% BODY TEXT
\section{Related Work}

The research in the Generative Deep Modelling can be broadly divided into 2 categories, engineering the loss function to guide the learning of the parameters for the generator and discriminator in the adversarial setting,  and the architecture design by exploring different architectures for generators and discriminators that can converge faster and produce high-quality images. In the architecture design, we can further divide the research into Generator Design and Discriminator Design. The Generator is the part of the GANs which is responsible for producing the images and learns the data distribution through the Discriminator with the help of a loss function, therefore the design of the loss function and Discriminator is pivotal, researchers \cite{karras2019style}explored the novel architectures for the Generator part of the GAN model to generate realistic and high-quality images. The discriminator usually is in the form of several convolutional layers, pretrained features and random projections were used in the Discriminator architecture by \cite{liu2020towards}.\cite{albuquerque2019multi} Multiple discriminators can be used against a single generator in the GANs training setting, these approaches are useful because they target a single consolidated optimization objective,\cite{neyshabur2017stabilizing} employed several discriminators by incorporating multi-objective optimization in the GANs training which led to increased performance in resolving mode collapse, and training stability, however, these approaches which utilize multiple discriminators are generally computationally expensive which leads to increasing the processing time for the GANs which can make them challenging to train and deploy. Depth wise convolutions have been studied and suggested to reduce the computation time for GANs\cite{ngxande2019depthwisegans}. Training the generator by incorporating multiscale feedback has been found to be useful in image synthesis \cite{karnewar2020msg} and can be used as a possible alternative to progressively growing techniques.\cite{schonfeld2020u} proposed an encoder-decoder architecture for the discriminator inspired by the image segmentation capacity of the U-Net to generate realistic images, by training the generator with pixel-wise feedback.Overfitting in the GANs can be minimized by using the  differentiation technique \cite{tran2021data,zhao2020differentiable}.Pre-trained models can be used to train the GANs, this is very useful because the pre-training can be done without the constraints of adversarial objectives in the pre-training phase. Pre-training provides the dis-entangled causal generative features which can be used to generate images using gradient ascent in the latent space learns the data distribution by training the generator with moment matching in the pre-trained network, but the results produced by this technique lag behind GANs in terms of data quality. One method frequently used is the addition of the losses, both the adversarial and perceptual losses are added \cite{nguyen2017plug,santos2019learning}. One method frequently used is the addition of the losses, both the adversarial and perceptual losses are added \cite{dosovitskiy2016generating,ledig2017photo}, which can only be implemented if the reconstruction targets are available as in the case of image to image translation.\cite{sauer2021projected}authors designed a novel architecture of the Discriminator with random projections to better utilize the deeper layers or the deep representations from the pre-trained model and feature pyramids to enable multi-scale feedback. The Projected GAN is fast in converging but comes at a cost of induced artifacts in the generated images even with the state-of-the-art FID scores. This research aims to solve this problem by experimenting and exploring the Generator design, the Generator used in the Projected GAN is based on the generator of the Fast GAN \cite{liu2020towards} which used skip connections powered by skip layer excitation instead of progressive growing GAN architecture of the Style GAN \cite{karras2019style}. We believe that the artifacts introduced in the images are produced in the low-resolution layers of the generator network which get propagated to the high-resolution layers due to up-sampling convolution layers, therefore by smoothing the lower resolution layers and avoiding the artifacts generation in them, the goal associated with the quality of the images to be realistic can be achieved, we propose that by combining the architecture strategies and advantages from Style GAN, Projected GAN, and  Fast GAN more realistic images can be generated with less training time, however, no single mixing strategy guarantees the optimal results therefore in our work we propose to incorporate the mapping network of the Style GAN into Fast GAN generator at low-resolution layers to avoid the artifacts propagation through skip layer excitation, which will lead to combining the advantages of high-quality image generation of Style GAN with the  Fast GAN.

%%%%%%%%% BODY TEXT

\section{Stylized Projected GANS}

GANS are trained to learn the data distribution of a dataset in an adversarial setting. It has 2 parts the Generator and the Discriminator. The Generator G takes the vector \textbf{z}, from a sample distribution \textbf{P(z)} and maps the latent vector to the generator space \textbf{G(z)}. The role of the discriminator is to accurately differentiate between the real samples \textbf{x} coming from the real data distribution \textbf{P(x)} and the fake sample \textbf{G(z)} generated by the generator. This leads to an adversarial game setting with the following loss for the GAN.\hfill \break
\begin{equation}
    min_{G} max_{D} \left ( E_{x}[log D(x)] + E_{z} [log (1- D(G(z)))]  \right ) 
\end{equation}

\hfill \break
SPGAN generator consists skip layer excitation, adaptive instance normalization and a mapping network.
\hfill \break
\subsection{Skip Layer Excitation}
The generator of a GAN model has an objective to generate realistic images, for the generation of the realistic images the generator usually consists of several convolution layers for up-sampling to create high-resolution images starting from low resolution. As the generator architecture becomes deeper the number of parameters to be optimized increases which leads to a higher training time for the generator, due to which the overall training time and computation resources needed to train a GAN model increases, one more factor contributing to the slow training is the weaker Gradient Flow through the successive layers of a deeper GAN network due to up-sampling \cite{karras2017progressive,zhang2017stackgan}.\cite{he2016deep} proposed a method for training deep neural networks with better signal flow between layers and called it residual learning. In residual learning, the layers learn the residual mapping instead of learning the input to output function mapping. Using the skip layer inputs from the previous layers and fitting a residual mapping it was shown that deeper neural networks can better optimize the parameters while overcoming the problems of saturation and accuracy degradation. Res-Blocks increase the computation cost of the training while increasing the signal flow. The Skip Layer Excitation module used in the SPGAN resembles the SLE from Fast GAN, the Res-Block consists of element-wise multiplication which leads to channel-wise multiplication of the activations, due to this it can be applied to two different resolution blocks. The SLE as defined in \cite{liu2020towards} can be stated as follows  \hfill\break
\begin{equation}\label{eq:2}
    {%
    y = F(x_{low,{W_i}}).x_{high}
        }
\end{equation}

\hfill\break
where ${x}$ is the input feature map, $y$ is the output feature map in the SLE block, the F corresponds to the functions and operations applied on the input feature maps, ${W_i}$  is the set of learn-able weights. The SLE is applied on the feature maps that are very different from each other in terms of resolution and is applied to the layers that are very far from each other in a deeper architecture, which leads to channel wise recalibration of the feature maps and improves the gradient flow of the model. The $\mathbf{x_{low}}$ can be thought of as the style attributes, by changing this input feature map the network generates an image consistent with the content but with the style of  $\mathbf{x_{low}}$.\hfill\break
\begin{table}[t]
  \centering
  \begin{tabular}{c|c}
    \toprule
    Model & FID \\
    \midrule
    Projected GAN & 29.50 \\
    Projected GAN + stylegan2 & 49.42 \\
    \bottomrule
  \end{tabular}
  \caption{Results on Pokemon dataset with 1000 images}
  \label{tab:pokemon1}
\end{table}

\begin{table}[t]
  \centering
  \begin{tabular}{c|c|c}
    \toprule
    Model & FID & Training\\
    \midrule
    Projected GAN & 5.86 & 25M \\
    SPGAN+AdaIN-8 & 5.55 & 23M \\
    SPGAN+AdaIN-16 & 4.67 & 20M \\
    SPGAN+AdaIN-32 & \textbf{3.98} & 17M \\
    \bottomrule
  \end{tabular}
  \caption{Results on FFHQ dataset with 52000 images}
  \label{tab:ffhq1}
\end{table}

\subsection{Mapping Network}
Inspired by the Style GAN architecture the SPGAN also uses a mapping network that takes an input of a random point from a latent space a generates a stylizing latent vector which is used as input to the successive layers by adaptive instance normalization AdaIn. The stylizing mapping network operates in parallel to the synthesis network with a function to control the style of the image. \hfill\break
\subsection{SPGAN Architecture}

SPGAN builds its architecture on the top of the Projected GAN, the discriminator design was the important contribution of the authors of the Projected GAN paper \cite{sauer2021projected}. In SPGAN the training is done similarly as in Projected GAN therefore the loss function remains similar to the original Projected GAN loss, in the case of the  projected GAN Discriminator,  the projections \textbf{Pj} are mapped from  the real and generated samples to discriminator's input space, therefore in case of Projected GANs the loss can be formulated as follows  \hfill \break

\begin{equation}\label{eq:3}
\resizebox{1\hsize}{!}{%
$ \min_{G} \max_{D_m} \sum_{m\in M}\left( E_{x}[ \log D_m(P_j(x))] + E_{z} [\log (1 - D_m(P_j(G(z))))] \right) $}
\end{equation}

% \begin{align}\label{eq:2}
%     min_{G} max_{D_m} \sum_{m\in M}^{}\left ( E_{x}[log Dm(Pj(x))] + E_{z} [log (1- D_m(Pj(G(z))))]  \right )
% \end{align}

\hfill \break
The ${D_m}$ in the Projected GAN loss refers to the independent discriminators, these independent discriminators individually operate on different projections of the features. The parameters for the ${Pj}$ are fixed because it corresponds to the pre-trained network responsible for producing the feature projections to be used by the discriminators. \hfill \break

The SPGAN Architecture is the combination of Fast GAN and Style GAN Generator however for the discriminator, Projected GAN is used. After running multiple combinations of various architecture mixing strategies in different data sets, we propose the SPGAN Generator architecture which is our main contribution to this project.\hfill \break

\begin{figure}[t]
    \centering
    \includegraphics[width=0.9\linewidth]{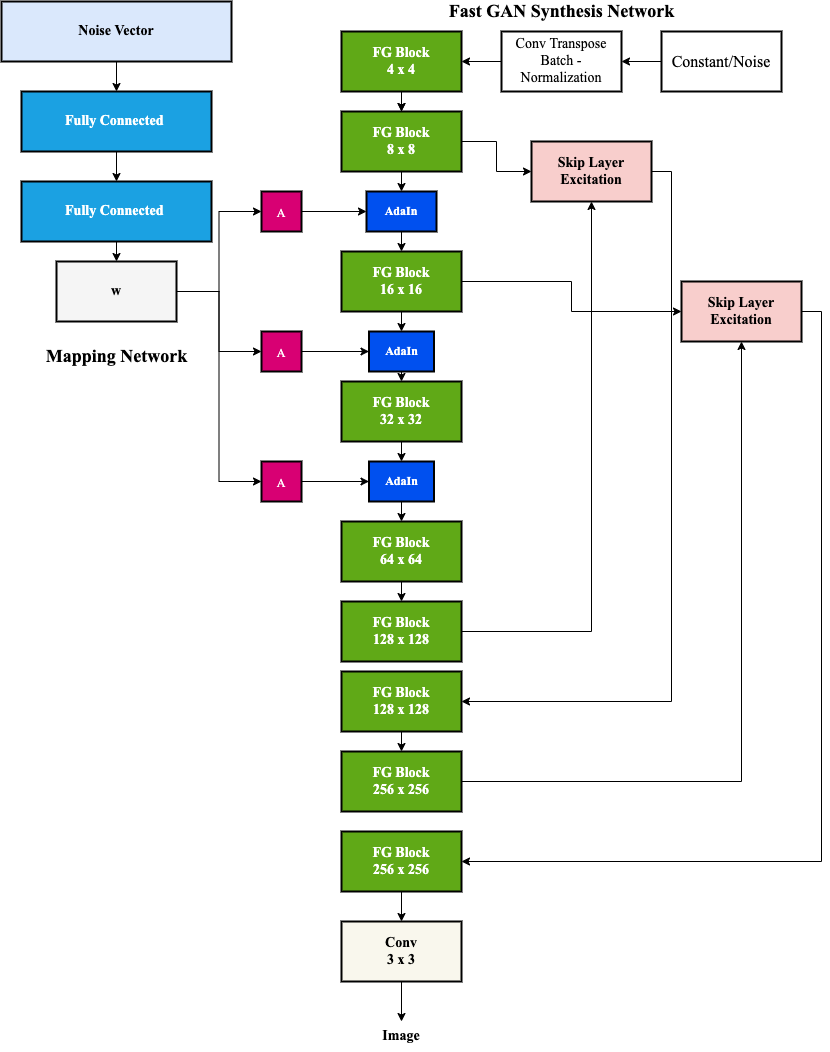}
    \caption{Proposed Generator Architecture for SPGAN: The SPGAN Generator consists of a Mapping Network and Skip Layer Excitation Modules across Up-Sampling Layers to generate images. There are two main differences with the previous Fastgan generator. Fast, the mapping network and Adaptive instance normalization. Second, The FG-BLOCK before each AdaIN layer is a lightweight block where we consider only upsampling, convolution, and noise injection. In Fast GAN, these blocks consist a batch-normalization which we removed with instance normalization}
    \label{fig:archi}
\end{figure}

Figure \ref{fig:archi} explains the Generator Architecture of the SPGAN. There are three main components of the SPGAN Generator, the Mapping Network, Fast GAN Blocks, AdaIn, and Skip Layer Excitation. The Mapping Network in this architecture consists of a fully connected Deep Neural Network which takes in the input of the noise vector and produces an output weight, these weights then pass through an Affine Transformation, this affine transformation is a learnable transformation consisting of a single layer neural network which matches the dimensions of the weights to the FG Block for AdaIn.The mapping network has been introduced in the architecture to resolve the artifacts produced in the images in the low-resolution FG blocks by introducing a learnable mapping network the artifacts generation can be minimized.  

\begin{figure}[t]
  \centering
%   \fbox{\rule{0pt}{2in} \rule{0.9\linewidth}{0pt}}
   \includegraphics[width=0.8\linewidth]{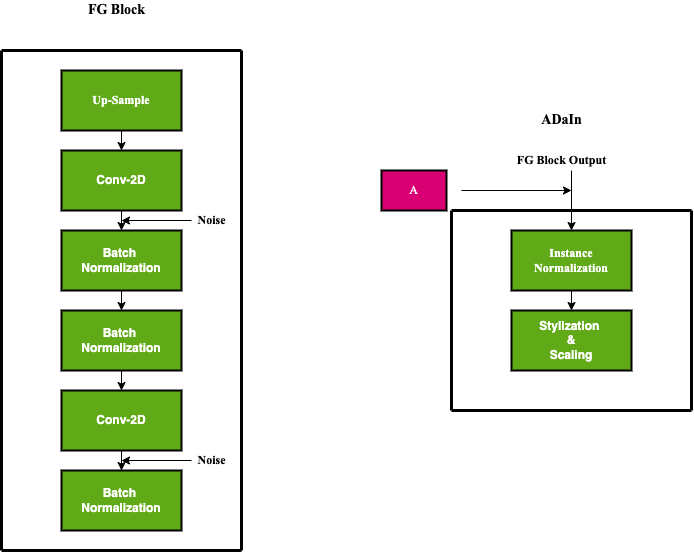}
   \caption{ Original FG-Block and AdaIN. This original FG block is used in the later layers where Skip layer excitation is used}
   \label{fig:onecol}
\end{figure}

\begin{figure}[t]
  \centering
%   \fbox{\rule{0pt}{2in} \rule{0.9\linewidth}{0pt}}
   \includegraphics[width=0.8\linewidth]{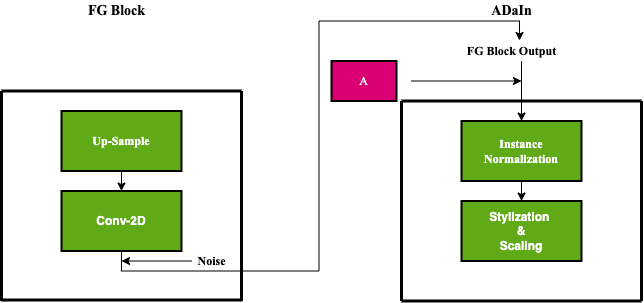}
   \caption{A light-weight FG-Block we propose and AdaIn block which we use in the first three layers of the SPGAN}
   \label{fig:onecol_1}
\end{figure}

\hfill \break
The affine transformation consists of a single hidden layer, which maps and reshapes the weights from the mapping network to the progressively growing Generator, mapping has been incorporated at the low-resolution levels because artifacts produced at the lower resolutions get amplified and propagated in the later higher resolutions.\hfill \break
FG block and AdaIn are shown in Figure \ref{fig:onecol} and \ref{fig:onecol_1}, FG block consists of multiple Up-Sampling, 2D Convolutions, and Batch Normalization layers. A single FG block contains an Up-Sampling layer followed by a 2D-Convolution Layer, the output of the convolution is mixed with the random noise and passed to a Batch Normalization Layer, after this layer the output flows to the 2D Convolution Layer, which adds noise to its output and passes to the batch normalization layer, which passes its output to the successive blocks.

\begin{figure}[t]
  \centering
%   \fbox{\rule{0pt}{2in} \rule{0.9\linewidth}{0pt}}
   \includegraphics[width=0.8\linewidth]{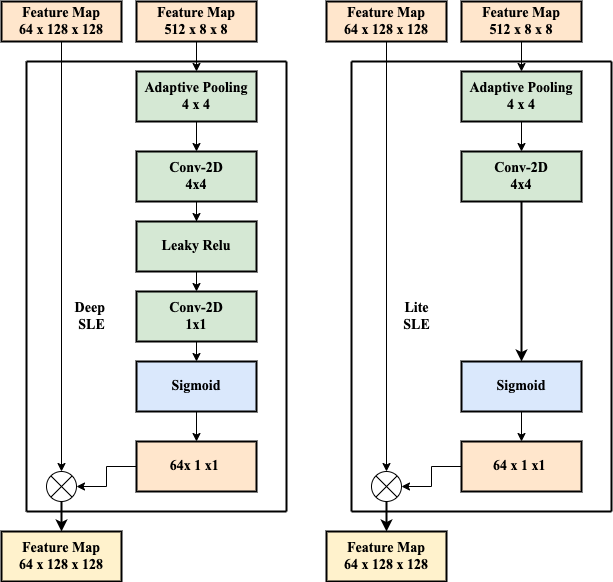}
   \caption{The Deep and the Lite SLE blocks}
   \label{fig:SLE blocks}
\end{figure}

% \begin{figure}
%     \centering
%     \includegraphics[width=0.4\textwidth, height=0.45\textwidth]{}
%     \caption{A light-weight FG-Block & AdaIn block which we use in the first three layers of the SPGAN}
%     \label{fig:modules}
% \end{figure}

\begin{table*}[t]
  \centering
  \begin{tabular}{c|c|c|c|c|c}
    \toprule
    Generator Model & FID ($\downarrow$) & KID ($\downarrow$)& Precision ($\uparrow$) & Recall ($\uparrow$) & Training ($\downarrow$)\\
    \midrule
    FastGAN &4.69 & 0.000858 & 0.69 & \textbf{0.46} & 43M \\
        StyledFastGAN with AdaIN in L1 &3.94&0.000748& \textbf{0.71} &0.40&10M \\
            StyledFastGAN with AdaIN in L1+L2 &4.09&\textbf{0.000598}& 0.69 &0.35&10M \\
    StyledFastGAN with AdaIN in L1+L2+L3  &\textbf{3.90}& 0.000659 &0.65&0.30 & 17M \\
    StyledFastGAN with AdaIN Combined with SLE &120.43& 0.02915 &0.91&0.00017 & 10M \\
    \bottomrule
  \end{tabular}
  \caption{Results on FFHQ dataset with 52000  images for generator search on different combinations of mapping network, AdaIN, SLE}
  \label{tab:ffhq52k}
\end{table*}

\begin{table*}[t]
  \centering
  \begin{tabular}{c|c|c|c|c|c}
    \toprule
    Generator Model & FID ($\downarrow$) & KID ($\downarrow$)& Precision ($\uparrow$) & Recall ($\uparrow$) & Training ($\downarrow$)\\
    \midrule
    FastGAN &4.26 & 0.000886 & 0.67 & 0.40 & 10M \\
        StyledFastGAN with AdaIN in L1 &3.73&0.000819&0.69 &\textbf{0.41}&10M \\
            StyledFastGAN with AdaIN in L1+L2 &\textbf{3.52}&\textbf{0.000646}& \textbf{ 0.68} &0.39&10M \\
    StyledFastGAN with AdaIN in L1+L2+L3  &4.13& 0.000650 &0.66&0.25 & 10M \\
    StyledFastGAN with AdaIN Combined with SLE &115.30&0.01856 &0.90&0.00000 & 10M \\
    \bottomrule
  \end{tabular}
  \caption{Results on FFHQ dataset with 70000  images for generator search on different combinations of mapping network, AdaIN, SLE}
  \label{tab:ffhq70k}
\end{table*}

\begin{table*}[t]
  \centering
  \begin{tabular}{c|c|c|c|c|c|c}
    \toprule
    Generator Model& Mapping Network & FID ($\downarrow$) & KID ($\downarrow$)& Precision ($\uparrow$) & Recall ($\uparrow$) & Training ($\downarrow$)\\
    \midrule
    StyledFastGAN with AdaIN in L1+L2+L3 & 2 Layers  &\textbf{4.13}& \textbf{0.000650} &0.66&0.25 & 10M \\
       StyledFastGAN with AdaIN in L1+L2+L3 & 4-Layers  &14.99& 0.00177 &\textbf{0.69}&0.19 & 10M \\
StyledFastGAN with AdaIN in L1+L2+L3 & 8-Layers  &5.35& 0.001054 &0.67&\textbf{0.34} & 10M \\
StyledFastGAN with AdaIN in L1+L2+L3 & 8-Layers  &5.06& 0.00119 &0.67&\textbf{0.35} & 43M \\
    \bottomrule
  \end{tabular}
  \caption{Results on FFHQ dataset with 70000  images for generator with deeper mapping networks}
  \label{tab:ffhq70k-deep}
\end{table*}

\begin{table*}[t]
  \centering
  \begin{tabular}{c|c|c|c|c|c|c|c}
    \toprule
    Generator Model& SLE &FID ($\downarrow$) & KID ($\downarrow$)& Precision ($\uparrow$) & Recall ($\uparrow$) & Training ($\downarrow$) & Dataset\\
    \midrule
    StyledFastGAN with AdaIN in L1+L2+L3 &Lite&4.42& 0.000703 &\textbf{0.67}&\textbf{0.31} & 10M & FFHQ-70K\\
        StyledFastGAN with AdaIN in L1+L2+L3 &Deep&\textbf{4.13}& \textbf{0.000650} &0.66&0.25 & 10M& FFHQ-70K\\
\midrule
      StyledFastGAN with AdaIN in L1+L2+L3 &Lite &4.41& 0.000897 &\textbf{0.69}&\textbf{0.33} & 10M & FFHQ-52K\\ 
            StyledFastGAN with AdaIN in L1+L2+L3 &Deep &\textbf{3.90}& \textbf{0.000659} &0.65&0.30 & 17M & FFHQ-52K\\ 

    \bottomrule
  \end{tabular}
  \caption{Results on lighter SLE block with Adain in the first  3 layers}
  \label{tab:ffhq-sle}
\end{table*}

\section{Experiments}
Numerous experiments were carried out using different architectures and mixing strategies with the goal of generating realistic images in less time, to decrease the learnable parameters of the model for faster and more realistic image generation. The experiments were carried out by focusing on the FFHQ dataset which is a big dataset for comparing various model architectures understudy and judging the performance of the proposed architecture.\hfill \break 
Experiments were carried out using a different mix of the architectures for the Generator part of the GAN model, these experiments mostly focus on the generator architectures and the associated experiments because the goal of this study is to minimize the artifacts generation observed in the Projected GAN when tested on the faces, the Projected GAN focuses on the discriminator design while using the Generator of the Fast GAN. The projected GAN discriminator is a novel architecture that used random projection and feature pyramids from the pre-trained model ( Efficient Net1)  by incorporating Cross Channel and Cross Scale Mixing, however, by using the pre-trained network there is a possibility that the projection space does not capture the some of the features in the new data set that the pre-trained network was not trained on. Therefore we divide our experiments into multiple groups for the Generator Search.  \hfill \break 
\subsection{Generator Search}
The experiments that were performed on the Generator part of the model with different combinations of Mapping Network, and  Slip Layer Excitation can be divided into two types, one in which we replaced the entire generator architecture with the Style GAN 2 generator and the second in which we mixed the mapping network from the Style GAN and Skip Layer Excitation from the FAST GAN generator.\hfill \break 
\subsubsection{Generator Swap}
In the first set of experiments, the two models were compared against each other on a small data set (Pokemon). The discriminator was not changed and the generator was swapped with the Style GAN 2 generator, because Style GAN 2 is known to produce very realistic images, we hypothesized that replacing the generator of the FAST GAN which was being originally used in Projected GAN might result in a better FID score and realistic images. The Style GAN generator is a progressive growing network with a mapping network and synthesis network, the mapping network comprising of 8 hidden layers and is focused on learning an affine transformation which is then normalized using adaptive instance normalization and fed to the subsequent layers.
\begin{equation}
\text{AdaIn} = \left(\sigma_{y} \frac{x - \mu_{x}}{\sigma_{x}} + \mu_{y}\right)
\end{equation}

where \begin{math}
{ \mu _{x},\mu _{y},\sigma _{x},\sigma _{y}}
\end{math} are the mean and standard deviation of the output from the mapping network and progressive growing GAN.
\hfill \break

The evaluation metrics used in this study include Frechet Inception Distance (FID), Kernel Inception Distance (KID), Precision and Recall.
 FID score is given by 

\begin{equation}
\text{FID} = \left | \mu_{g} - \mu_{r} \right |^{2} + \text{tr}\left[{\Sigma_{g}} + {\Sigma_{r}} + 2\left({\Sigma_{g}}^{1/2}{\Sigma_{r}}{\Sigma_{g}}^{1/2}\right)^{1/2}\right]
\end{equation}

where \begin{math}
{ \mu _{g},\mu _{r},\Sigma _{r},\Sigma _{g}}
\end{math} represent  the feature wise average and covariance of the generated samples and the real images respectively.\hfill\break
Apart from FID, KID, Precision, and Recall are used to evaluate the model performance in different cases.KID measures the dissimilarity between two probability distributions by using independent samples from the distributions to be compared, KID is an unbiased alternative metric for FID. Precision indicates the quality of the generated images and Recall indicates the class or sample diversity. It is observed that GANS produce images with high precision and relatively low recall due to mode collapse.

By replacing the Generator of the Style GAN we tested our model against the metric of FID Score.
The results of our initial experiment are shown in the following Table \ref{tab:pokemon1}, we tested our model based on the Generator of the Style GAN 2 and the discriminator of the Projected GAN, and the original Projected GAN, which uses a novel architecture for the discriminator while the generator of the FAST GAN, the data-set used is Pokémon Data Set ( 833 images ), the reason for selecting the few numbers of images is based on the results from the Projected GAN paper that Projected GANS converge faster. The initial results turned out to be surprising, as we believed that the FID Score shall improve when we used the Style GAN progressive growing Generator but to our surprise, the FID score has significantly increased as indicated in the Table
\ref{tab:pokemon1}\hfill\break

\subsubsection{Mapping Network Integration with Different Resolutions}
The mapping network was integrated with the different layers of the generator as shown in Figure \ref{fig:archi}. These experiments on SPGAN with stylization in the different number of layers were carried out with the purpose to observe the effect of stylization at earlier and later layers in the synthesis network. Experiments with the stylization at the resolution of 8x8,16x16 and 32x32 were performed, integration with the deeper layers are also performed and the results are given in Table \ref{tab:ffhq1}. In this set of experiments we considered the FFHQ data set which is a big data set, to understand the artifact generation process and to minimize the production of the artifact, the FAST GAN generator block has also been altered to lower the complexity and reduce the training parameters, the  Figure \ref{fig:onecol} shows the original FG Generator Block where Upsampling is followed by sequential Conv-2D, Batch Normalization layers multiple times whereas in the lighter architecture the Upsampling is followed by a single layer of Conv-2D which is passed to the AdaIN block along with the affine transformation from the mapping network, which undergoes Instance Normalization as shown in the Figure \ref{fig:onecol_1}.  
\hfill\break

\subsubsection{Deeper Mapping Network}
It has been observed that deep neural networks are good at learning the non-linear relationships and generalizing to the data, the mapping network is a DNN, after the initial set of experiments with a shallow network a series of experiments were designed to test the models on different depths of the mapping network at 2, 4 and 8 layers, the details of which can be seen in Table \ref{tab:ffhq70k-deep}.\hfill\break

\subsubsection{Integrating Mapping Network and SLE}
As suggested in \cite{sauer2021projected}, the artifacts observed in the images generated by Projected GAN might be mitigated by combining the mapping network of the Style GAN and the Skip Layer Excitation from the FAST GAN. We designed our experiments by integrating the mapping network along with adaptive instance normalization and SLE. A series of experiments with different combinations of the mapping network networks and the synthesis network were carried out, for the first set of experiments the SLE is used along with the mapping network at different resolution levels, and for the second set the mapping network along with two variants of SLE block namely light and deep are tested, for the lite block the relu activation is removed, for the deep block for the SLE used is the same used in FAST GAN the SLE blocks are shown in the Figure \ref{fig:SLE blocks}, the details of experiments carried out are given in the Table \ref{tab:ffhq-sle},\ref{tab:ffhq52k}, and \ref{tab:ffhq70k}.\hfill\break

\section{Discussion \& Results}
Results obtained after the experimentation performed for Generator Search are listed in this section. FID, KID, Precision, and Recall are compared for the different mixing strategies on the FFHQ dataset. \hfill\break
\subsection{Generator Search Results}
Experimentation with different mixes for the generator design was performed as described in the previous section. The strategies used included Generator Swap, Mapping Network Integration into FAST GAN architecture at different resolution levels, testing for the effectiveness of various depths of the Mapping Network, and Integration of SLE and Mapping Network. \hfill\break
\subsubsection{Generator Swap}
The first set of experiments started with the simplest procedure of generator swap, in this experiment the generators of Fast GAN and Style GAN2, however, the discriminator was held constant to be Projected GAN, and the swapped architectures were compared on a small data set of Pokemon, the results of the experiment are given in \ref{tab:pokemon1}. With a simple swap, we observe that the FID score obtained with the Style GAN2 generator which is famous for generating realistic images is very high as compared to the Projected GAN which uses a Fast GAN generator. One reason can be that Style GAN2 is based on progressive growing GANS and relies on the discriminator's signal for generating realistic images therefore the discriminator of the Projected GAN is not providing a strong signal to the generator to train, also the size of the Pokemon dataset is very small and progressive growing GANs like Style GAN requires larger data sets to generalize to the underlying data distribution, therefore the FID scores observed with the Style GAN2 generator is very high.\hfill\break

\subsubsection{Mapping Network Integration at Different Resolutions}
Table \ref{tab:ffhq1} shows our results on the FFHQ dataset with resolution 256x256. Interestingly, with light-weight FG-BLOCK before each AdaIN layer, we can reduce the number of trainable parameters to half of the original FAST GAN. Thus, we lower training time with better FID scores. We can easily observe that our best SPGAN achieves an FID of 3.98 with 17M training samples whereas Projected GAN with FastGAN achieves an FID of 5.86 with 25M images. These results are different from the results reported in the Projected GAN paper as our FFHQ dataset is different where we have 52000 images. The results for stylization in different layers with 52,000 and 70,000 FFHQ data set is given in Table \ref{tab:ffhq52k} and \ref{tab:ffhq70k} , it can be observed that stylization and SLE do not work together in later and deep layers, for the 52k data set the SPGAN beats state of the art Projected GAN with Fast GAN generator, in FID, KID and precision and required 4X less samples. To ensure a balanced comparison the networks were  trained on 70K FFHQ data set, where each image is shown 10M times,it was observed that now SPGAN beats Projected GAN in every metric on same number of samples\hfill\break

\subsubsection{Deeper Mapping Network}
Mapping Networks with different depths were investigated for the performance against FID and KID scores on the FFHQ dataset. The results are listed in Table \ref{tab:ffhq70k-deep}. The lowest FID of 4.13 and KID of 0.00065 are achieved with a two-layer deep mapping network, however, the recall here is lower as compared to the 8 Layers Deep Mapping Network, highest precision of 0.69 with the lowest recall and highest FID and KID are seen for the 4-layers Deep Network. Therefore we observe that as the mapping network depth increases we do not see a significant improvement in precision which indicates the image quality relatively remains the same, however, recall increases which indicates mitigating the mode collapse problem. 8 Layers Deep Mapping network shows an overall optimal performance on all of the contrasting metrics with an FID score of 5.35, KID score of 0.0010, Precision of 0.67, and Recall of 0.34 on 10M images, training the same architecture up to 43M images does lower the FID score but does not significantly improve the performance on other metrics.  \hfill\break

\subsubsection{Integrating Mapping Network and SLE}
Two architectures of the SLE the lite and the deep are tested for the performance, the results are given in the Table \ref{tab:ffhq-sle}, with the lighter architecture the precision and recall are high, with a significant improvement observed in recall where the recall increased from 0.25 to 0.31 indicating an increase in model diversity, however, the precision remains very close for the Generator with 70k images for the 52K images recall and precision both increase for the lite architecture however the best FID and KID scores are achieved for a deep block on both 52K and 70K images.

\section{Conclusion}
SPGAN is the integration of stylization from Style GAN and SLE from Fast GAN, after performing numerous experiments and ablation studies with different architectures and mixing strategies. The different mixing strategies produce different FID, KID, precision, and recall scores, and we need to find a compromise between precision and recall. It was observed that lighter SLE blocks generate images with higher precision at a comparable FID score as compared to original FG SLE blocks. The depth of the mapping network was found to be linked with the image quality and mode collapse through precision and recall, shallow mapping networks with 2 hidden layers produce high-quality images but struggle at recall, however, the deeper mapping networks produce diverse images at a cost of a little reduction in precision. SPGAN with stylization in initial layers produces much better results as compared to deeper layers, which is an indication that artifacts are produced at low resolution and can be mitigated by mapping network to reduce their propagation to higher resolutions. Mapping networks along with SLE at the later layers produce poor FID, and precision scores. SPGAN with stylization in initial layers with lighter SLE blocks beats the state of the art Projected GAN in FID, KID, and precision with at least 4X fewer samples which makes SPGAN a sample efficient architecture to generate high-quality images fast.

\section{Future Works}
SPGAN readily reduces the number of samples required to generate high-quality images as compared to state-of-the-art Projected GAN for quick high-quality image generation, however, we still observe some artifacts in the produced images. We propose after conducting experiments, that the discriminator part of the model shall also be changed, one direction to proceed in this regard can be a loss function which takes into account the artifacts generation, another option can be a separate head that classifies an image as either with or without artifact, one more option can be to train an encoder which takes an input of the generated image, the activations are taken and clustering is performed, the number of clusters can then be an indication of the presence or absence of the feature in faces. 
% \usepackage{booktabs}

%\section{Work Division}
%Brain Storming Sessions, were done meeting was performed every week for background study, literature review, architecture design and design of experiments for the project, these steps were done together by both members. The code section was divided and experiment load was equally divided between the 2 members. The report was also equally divided for ensuring equal effort in the project, the details are given in \ref{tab:work-division}

%\begin{table*}[t]
%  \centering
%  \begin{tabular}{c|c}
%    \toprule
%    Task & Contributor \\
%    \midrule
%     Background study & Both  \\
%     Literature Review & Both\\
%     Architecture Design & Both\\
%        Experiment Design & Both \\
%        Coding & Both\\
%        Experiment on Pokemon &  Both\\
%        Experiment on FFHQ 52K & Malik Shaid Sultan %\\
%        Experiment on FFHQ 70K & Md Nurul Muttakin \\
%      Experiment on Deeper Mapping Network & Malik %Shahid Sultan \\
%      Experiment on lighter SLE block & Md Nurul %Muttakin \\
%      (Report) Abstract & Both \\
%      (Report) Introduction & Md Nurul Muttakin \\
%      (Report) Related Works & Malik Shahid Sultan \\
%      (Report) Design and Experiment Explanation & Both \\
%      (Report) Discussion \& Results  & Both\\
%      (Report) Conclusion   & Both\\
%      (Report) Future Works   & Both\\
%     
%    \bottomrule
%  \end{tabular}
%  \caption{Work Division}
%  \label{tab:work-division}
%\end{table*}

%%%%%%%%% REFERENCES
{\small
\bibliographystyle{egbib}
\bibliography{egbib}
}

\end{document}